\pdfoutput=1
\documentclass[11pt]{article}

\usepackage[preprint]{acl}

\usepackage{times}
\usepackage{latexsym}

\usepackage[T1]{fontenc}
\usepackage[utf8]{inputenc}

\usepackage{microtype}

\usepackage{inconsolata}

\usepackage{graphicx}

\usepackage{amsmath}
\usepackage{multirow}
\usepackage{graphicx}
\usepackage{wrapfig}
\usepackage{booktabs}
\usepackage{amssymb}
\usepackage[most]{tcolorbox}
\usepackage{xcolor}
\usepackage{colortbl}
\usepackage{array}
\usepackage{color-edits}
\usepackage{soul}
\usepackage{pifont}

\newcommand{\think}[1]{\textcolor{gray}{\texttt{<think>}} #1 \textcolor{gray}
{\texttt{</think>}}}
\newcommand{\evi}[1]{\textcolor{blue}{\texttt{<original\_evidence>}} #1 \textcolor{blue}{\texttt{</original\_evidence>}}}
\newcommand{\search}[1]{\textcolor{cyan}{\texttt{<search>}} #1 \textcolor{cyan}{\texttt{</search>}}}
\newcommand{\obs}[1]{\textcolor{brown}{\texttt{<observation>}} #1 \textcolor{brown}{\texttt{</observation>}}}
\newcommand{\answer}[1]{\textcolor{purple}{\texttt{<answer>}} #1 \textcolor{purple}{\texttt{</answer>}}}

\newcommand{\hevi}{\textcolor{blue}{\texttt{<original\_evidence>}}}
\newcommand{\hsearch}{\textcolor{cyan}{\texttt{<search>}}}
\newcommand{\hobs}{\textcolor{brown}{\texttt{<observation>}}}
\newcommand{\hanswer}{\textcolor{purple}{\texttt{<answer>}}}

\definecolor{lightgreen}{rgb}{0.6, 1, 0.6}
\definecolor{lightred}{rgb}{1, 0.6, 0.6}
\definecolor{lightyellow}{rgb}{1, 1, 0.6}
\definecolor{lightblue}{rgb}{0.6, 0.8, 1}
\definecolor{lightgrey}{rgb}{0.93, 0.93, 0.93}
\definecolor{myred}{rgb}{0.7, 0.3, 0.0}
\definecolor{myblue}{rgb}{0.2, 0.3, 0.6}
\definecolor{mypurple}{rgb}{0.5, 0, 0.5}
\definecolor{mygreen}{rgb}{0.0, 0.4, 0.2}
\definecolor{mybrown}{rgb}{0.65, 0.16, 0.16}
\definecolor{mygray}{rgb}{0.5, 0.5, 0.5}
\definecolor{relcolor}{rgb}{1.0, 0.7, 0.4}

\newcommand{\hlrel}[1]{{\sethlcolor{relcolor}\hl{#1}}}

\title{R-Search: Empowering LLM Reasoning with Search\\
via Multi-Reward Reinforcement Learning}

\author{
 Qingfei Zhao$^{1,2}$,
 Ruobing Wang$^{1,2}$,
 Dingling Xu$^{3}$,
 Daren Zha$^{1*}$,
 Limin Liu$^{1}$
\\
\\
 \textsuperscript{1}Institute of Information Engineering,
Chinese Academy of Sciences\\
 \textsuperscript{2}School of Cyber Security,
University of Chinese Academy of Sciences\\
 \textsuperscript{3}Beijing Normal University\\
 \texttt{\{zhaoqingfei,wangruobing,zhadaren\}@iie.ac.cn}
}

\begin{document}
\maketitle
\renewcommand{\thefootnote}{\fnsymbol{footnote}}
    \footnotetext[1]{Corresponding author
    }
\begin{abstract}
Large language models (LLMs)
have notably progressed in multi-step and long-chain reasoning. However, extending their reasoning capabilities to encompass deep interactions with search remains a non-trivial challenge, as models often fail to identify optimal reasoning–search interaction trajectories, resulting in suboptimal responses.
We propose \textbf{R-Search}, a novel reinforcement learning  framework for \textbf{R}easoning–\textbf{Search} integration, designed to enable LLMs to autonomously execute multi-step reasoning with deep search interaction, and 
% capable of learning
learn optimal reasoning–search interaction trajectories via multi-reward signals, improving response quality in complex logic- and knowledge-intensive tasks.
R-Search guides the LLM to dynamically decide when to retrieve or reason, while globally integrating key evidence to enhance deep knowledge interaction between reasoning and search.
During RL training, R-Search provides multi-stage, multi-type rewards
to jointly optimize the reasoning–search trajectory.
Experiments on seven datasets show that R-Search
outperforms advanced RAG baselines by up to 32.2\% (in-domain)
and 25.1\% (out-of-domain).
The code and data are available at \url{https://github.com/QingFei1/R-Search}.
\end{abstract}

\section{Introduction}

Large language models (LLMs) have demonstrated substantial progress across a wide range of natural language processing (NLP) tasks, driven by their impressive language understanding and reasoning abilities
\cite{DBLP:journals/corr/abs-2303-08774, DBLP:journals/corr/abs-2407-11511}.
In logic-intensive tasks~\cite{DBLP:conf/acl/AsaiH20,shao2024deepseekmath}, state-of-the-art LLMs, exemplified by DeepSeek-R1 \cite{DBLP:journals/corr/abs-2501-12948}, have demonstrated remarkable capabilities in long-chain and multi-step reasoning
\cite{DBLP:journals/corr/abs-2412-16720}.
In knowledge-intensive tasks~\cite{DBLP:conf/nips/LewisPPPKGKLYR020,DBLP:conf/acl/TrivediBKS23}, even LLMs with strong reasoning capabilities are susceptible to generating hallucinated outputs \cite{DBLP:journals/corr/abs-2309-01219}.
This primarily arises from inherent limitations in the accuracy, timeliness, and coverage of their parametric knowledge.
To mitigate hallucination, LLM-based Retrieval-Augmented Generation (RAG) \cite{DBLP:conf/nips/LewisPPPKGKLYR020, DBLP:conf/emnlp/ZhaoWCZTD024} incorporates search actions before generation, enabling the LLM to augment its input with non-parametric knowledge in textual form. This allows the LLM to flexibly access and integrate relevant information from external knowledge sources, thereby enhancing the reliability of downstream generation.
However, downstream generation often struggles to benefit from reasoning alone or one-time search when addressing more complex logic- and knowledge-intensive tasks, e.g., multi-hop question-answering (QA) task~\cite{DBLP:conf/emnlp/Yang0ZBCSM18}.
In tackling such complex tasks, the LLM is expected to dynamically integrate external knowledge into the reasoning process, not only to bridge the knowledge gap but to guide and deepen the reasoning trajectory.

Previous multi-turn RAG methods~\cite{DBLP:conf/naacl/JeongBCHP24, DBLP:conf/acl/TrivediBKS23} enable the integration of external knowledge into reasoning by prompting LLMs to iteratively perform reasoning–search interactions. In this process, retrieved information enhances the model’s reasoning, which in turn guides the subsequent retrieval, forming a dynamic loop between them. However, these methods typically rely on the LLM’s internal cognition to decide when and what to retrieve, leading to two main limitations:
\textbf{1)} The retrieval timing determined by the model’s internal knowledge distribution does not always align with the actual need for retrieval; \textbf{2)} the modular and decoupled design of reasoning and search limits deep interaction of external knowledge into the reasoning chain. As a result, the model often makes decisions based only on partial information from previous searches or thoughts.
These limitations lead to suboptimal or even incorrect reasoning–search trajectories, ultimately reducing the quality of the final outputs.

To this end, we propose R-Search, a novel reinforcement learning (RL)-based framework that enables LLMs to dynamically interleave multi-step reasoning and search, and to learn optimal reasoning–search trajectories through multi-reward signals.
\textbf{First}, R-Search allows the LLM to trigger retrieval at any token-level reasoning step, seamlessly integrating retrieved content into the reasoning process for deeper coupling between reasoning and external knowledge.
After the interaction, the LLM distills retrieved documents into evidence through reasoning.
This facilitates the LLM in re-evaluating and structuring critical knowledge from a global perspective, thereby enhancing its focus on the facts most pertinent to solving the task.
\textbf{Second,} we design a multi-stage, multi-type reward mechanism that incorporates answer quality, evidence quality, and format correctness as reward signals. These complementary signals promote the model to learn the optimal reasoning–search interaction sequence. In particular, the evidence reward encourages the model to focus on the factual quality of key intermediate reasoning steps, promoting more robust reasoning paths and reducing the risk of shortcut-driven or speculative behavior.

Our contributions are threefold:
\textbf{1) Framework Design:}
we propose R-Search, a novel RL-based framework that jointly optimizes complex reasoning–search trajectories in RAG. R-Search promotes robust policy learning by interleaving multi-step reasoning with dynamic search and optimizing through multi-reward modeling. It effectively guides the LLM to ensure both the soundness of intermediate reasoning and the completeness of retrieved knowledge.
\textbf{2) Superior Performance and Insightful Analysis:} we conduct extensive experiments on seven datasets across both multi-hop and single-hop QA tasks, demonstrating the superiority of R-Search over vanilla and advanced RAG baselines (up to 32.2\%). Further analyses—including ablation and training dynamics—validate the effectiveness of evidence integration and multi-reward modeling, and provide insights into performance trends and retrieval behaviors under different RL algorithms.
\textbf{3) R-Search-as-a-Tool:} 
  % 在此填写第三点内容
    We propose R-Search-as-a-Tool (RSTool), which modularizes high-quality evidence in reasoning into transferable components, enabling the offloading of complex and costly reasoning–search interactions to local deployments, with strong practical scalability.

\section{Related Work}
\subsection{Retrieval-Augmented Generation (RAG)}
RAG typically follows a retrieve-and-generate paradigm~\cite{DBLP:conf/nips/LewisPPPKGKLYR020, DBLP:journals/corr/abs-2502-01142}.
The retrieval corresponds to the Search action, which involves acquiring external non-parametric knowledge from various sources using different search tools.
The generation refers to the tokens produced by LLM reasoning, encompassing both the intermediate reasoning process and the execution of specific modules in modular RAG, including the final answer generation~\cite{DBLP:journals/corr/abs-2312-10997}.
Recently, LLM-based RAG systems~\cite{DBLP:journals/tacl/RamLDMSLS23, DBLP:conf/iclr/YoranWRB24} have demonstrated significant performance gains across various NLP tasks~\cite{DBLP:conf/nips/BrownMRSKDNSSAA20, DBLP:journals/corr/abs-2303-08774, DBLP:journals/corr/abs-2407-11511}, especially in open-domain question answering.
To produce higher quality responses, some branching RAG~\cite{DBLP:conf/iclr/KimNMP0S0S24, DBLP:conf/naacl/ShiMYS0LZY24} methods summarize the retrieved documents separately for multiple candidate responses to improve understanding of external knowledge.
However, as tasks involving more complex reasoning (e.g. multi-hop QA tasks~\cite{DBLP:conf/emnlp/Yang0ZBCSM18, DBLP:conf/coling/HoNSA20, DBLP:journals/tacl/TrivediBKS22, DBLP:conf/emnlp/PressZMSSL23}) have emerged, traditional RAG methods~\cite{DBLP:journals/corr/abs-2002-08909, DBLP:conf/nips/SachanRHDY21} struggle with insufficient external knowledge integration. Several advanced RAG methods~\cite{DBLP:conf/naacl/JeongBCHP24, DBLP:conf/emnlp/JiangXGSLDYCN23, DBLP:conf/emnlp/ChengLLZYSLS0Q24, DBLP:conf/acl/TrivediBKS23, DBLP:conf/emnlp/ShaoGSHDC23, DBLP:journals/corr/abs-2404-00610, DBLP:conf/iclr/AsaiWWSH24} have attempted multi-turn reasoning–search interactions to enable deeper knowledge exploration, including multi-step~\cite{DBLP:conf/acl/TrivediBKS23, DBLP:journals/corr/abs-2404-00610, DBLP:conf/emnlp/ShaoGSHDC23} and adaptive RAG methods~\cite{DBLP:conf/naacl/JeongBCHP24, DBLP:conf/emnlp/JiangXGSLDYCN23, DBLP:conf/iclr/AsaiWWSH24}.
Nevertheless, these methods still heavily rely on carefully crafted prompts, making them difficult to scale and limiting the depth of interaction between search and reasoning.
This often results in suboptimal interaction.
In our work, we aim to construct an agent-based RAG pipeline that supports flexible reasoning–search interaction, and optimizes complex interaction trajectories via RL.
\begin{figure*}
  \centering
\includegraphics[width=\textwidth]{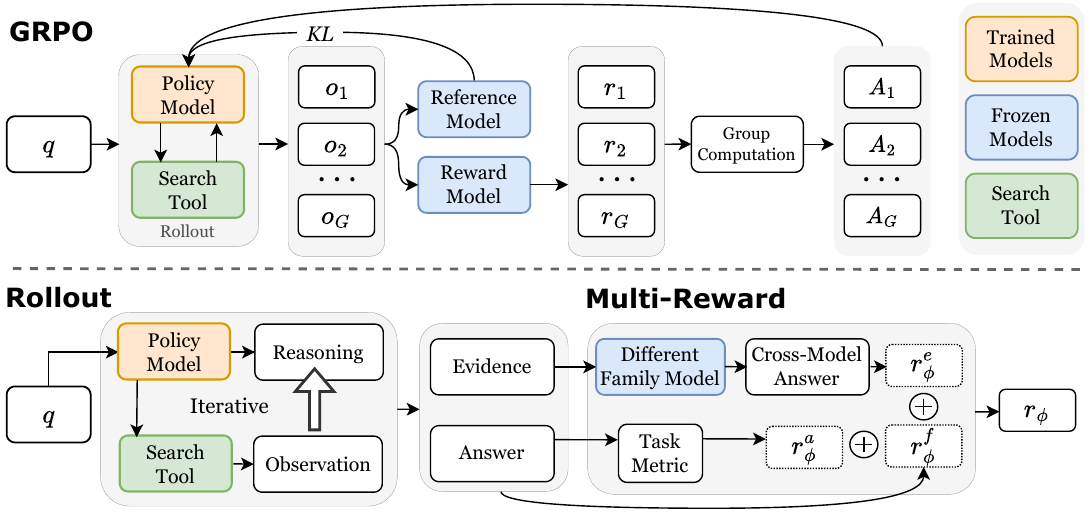}
  \caption{\textbf{Overview of R-Search.} }
  \label{fig1: Overview of R-Search}
\end{figure*}
\subsection{Reinforcement Learning for RAG}
Reinforcement learning (RL)~\cite{kaelbling1996reinforcement,wiering2012reinforcement} is an effective paradigm for enhancing the reasoning capabilities of LLMs. Recent studies~\cite{guo2025deepseek,shao2024deepseekmath,DBLP:journals/corr/abs-2412-16720} have shown that RL with rule-based reward functions enables models to acquire complex task reasoning and self-correction abilities~\cite{weng2022large,kumar2024training} without explicit intermediate supervision.
However, most existing RL approaches focus mainly on internal reasoning, with limited integration of external search information and insufficient handling of multi-turn interactions between reasoning and search.
Concurrent work, such as Search-R1~\cite{jin2025search}, applies RL algorithms, including Proximal Policy Optimization (PPO)~\cite{schulman2017proximal} and Group Relative Policy Optimization (GRPO)~\cite{shao2024deepseekmath}, to improve LLMs’ autonomous search and reasoning abilities.
Yet challenges remain in reward design and reasoning–search interaction.
In our work, we further explore RL-based RAG methods to enhance LLMs’ autonomous exploration, robust reasoning, and deep interaction with search.
\section{R-Search}
In this section, we introduce RL tailored for RAG  pipeline (\S~\ref{subsec: RL on RAG Paradigms}) and describe the training process (\S~\ref{subsec: R-Search Training}) of R-Search, as illustrated in Figure~\ref{fig1: Overview of R-Search}.
\subsection{RL on RAG Paradigms}
\label{subsec: RL on RAG Paradigms}
Advanced RAG tightly interweaves reasoning and search through several LLM-based components, enabling an adaptive and active process for exploring external knowledge to solve complex questions.
In this process, the LLM can determine its next action based on its current observation of the environment, such as the reasoning trajectory and the retrieved external knowledge observed so far. The possible actions include initiating a new search, continuing the reasoning process, or generating the final answer.
We formalize this advanced RAG paradigm as follows:
\begin{equation}
a_1 \rightsquigarrow a_2 \rightsquigarrow \cdots \rightsquigarrow a_T, \quad a_t \in \{ \mathcal{S}, \mathcal{R} \},
\end{equation}
where $\mathcal{S}$ and $\mathcal{R}$ denote the search and reasoning actions $a$, respectively, and $T$ is the total number of decision time steps in the action sequence.

Such a complex RAG pipeline can be viewed as a partially observable Markov decision process (POMDP).
Therefore, we develop a novel RL-based framework that optimizes the multi-turn interleaved trajectories of reasoning and search.
During training, we optimize the policy model by maximizing the following objective:
\begin{multline}
% \max_{\pi_{\theta}} 
\mathcal{J}(\theta ) =
\mathbb{E}_{q\sim \mathcal{D},o\sim \pi_{\theta}(\cdot \mid q;\mathcal{S})}\left [ r_{\phi }(q,o) \right ]\\
-\beta \mathbb{D}_{\text{KL}}[\pi_{\theta}(o \mid q;\mathcal{S})]\parallel \pi_{\text{ref}}(o \mid q;\mathcal{S})],
\label{eq:RL-objective}
\end{multline}
where $\pi_\theta$ is the policy model, $q$ and $o$ are the input question and generated output, and $\mathcal{S}$ denotes search mechanism. $r_\phi$ represents the reward function, $\pi_{\text{ref}}$ is the frozen reference model, and $\beta$ is a coefficient balancing the KL penalty.
Unlike $\pi_{\theta}(\cdot \mid q)$, the policy $\pi_{\theta}(\cdot \mid q; \mathcal{S})$ executes an interleaved process of reasoning-search to generate the rollout sequence, denoted as \textbf{\texttt{Reason} $\bowtie$ \texttt{Search}}.

\subsection{R-Search Training}
\label{subsec: R-Search Training}
\subsubsection{Rollout: evidence-augmented iterative reasoning and search}
\label{subsubsec: Rollout}
Table~\ref{system template} describes the system template used for rollout (more details in Appendix~\ref{subsec: System Template}).
We first prompt the LLM to generate a long Chain-of-Thought (CoT) based on the original question $q$, thereby constructing an explicit reasoning process.
During the reasoning process, we encourage the model to trigger the search action at appropriate points to acquire external non-parametric information. Whenever the model determines that search is needed at the current reasoning state, it generates a new search query enclosed within \search{and} tags.
We identify this specific search token and feed the generated query $q^{*}$ into a search tool to retrieve top-$k$ relevant documents $D_{k}=d_1,d_2,\cdots,d_k$. These documents are wrapped with specific tokens \obs{and}, and appended to the existing reasoning trajectory. The LLM then re-engages the reasoning process, ultimately forming an interactive loop between reasoning and search.
Next, when the LLM determines that the current state is sufficiently informative to produce a final answer $\alpha$, we prompt it to rethink all previously retrieved information and derive factual evidence $e$ that supports question resolution.
This enables the LLM to reason from a global perspective, leveraging all potentially relevant knowledge observed from the external information environment to support answer generation.
Moreover, we leverage the model’s internal reasoning capabilities to interpret and integrate external knowledge during the generation of factual evidence.
We wrap the factual evidence with the special tokens box \evi{and}.
After the evidence reasoning process, the LLM continues to generate the final answer $\alpha$, enclosed within the specific tokens box \answer{and}.
\begin{table}[ht]
\centering
    \renewcommand{\arraystretch}{0.8}
\small
\begin{tabular}{p{7.2cm}}
\toprule
\toprule
You are a helpful assistant that can solve the given question step by step. For each step, start by explaining your thought process. If additional information is needed, provide a specific query enclosed in \search{and}. The system will return the top search results within \obs{and}. You can perform multiple searches as needed.
When you know the final answer, use \evi{and} to provide all potentially relevant original information from the observations. Ensure the information is complete and preserves the original wording without modification. If no searches were conducted or observations were made, omit the evidence section. Finally, provide the final answer within \answer{and} tags.
\\
\bottomrule
\bottomrule
\end{tabular}
\caption{\textbf{System template.}
The question is appended at the end during training and inference.}
\label{system template}
\end{table}

\subsubsection{Multi-reward modeling}
\label{subsubsec: Multi-Reward Modeling}
Due to the high cost and potential bias associated with training reward models using human feedback, we follow~\cite{NEURIPS2024_c4e380fb}
and adopt the rule-based reward (RBR).
Considering the complementary effects of multiple rewards~\cite{dann2023reinforcement},
we design multi-dimensional, multi-stage reward signals $r_{\phi}$, including evidence rewards $r^{e}_{\phi}$, answer rewards $r^{\alpha}_{\phi}$, and format rewards $r^{f}_{\phi}$.
\paragraph{Answer reward.}
Metrics such as F1-score and EM are widely used to evaluate the correctness of model outputs.
We choose the moderately strict F1-score to construct the answer reward signal. Specifically, we extract the content $\alpha_{\text{pred}}$ within \answer{and} generated by the $\pi_{\theta}$ and compute the F1-score against the gold answer $\alpha_{\text{gold}}$, which serves as the answer reward $r^{\alpha}_{\phi}$.
\begin{align}
r^{\alpha}_{\phi}(q,o) &= \text{F1}(\alpha_{\text{pred}},\alpha_{\text{gold}}) \notag\\
&= \frac{2\cdot|\alpha_{\text{pred}}\cap \alpha_{\text{gold}}| }{|\alpha_{\text{pred}}|+|\alpha_{\text{gold}}|},
\end{align}
where $|\alpha_{\text{pred}} \cap \alpha_{\text{gold}}|$ is the number of word-level overlaps between the predicted and gold answers.

\paragraph{Evidence reward.}
In RAG systems, the quality of evidence directly impacts answer accuracy.
we introduce models from a different family with distinct policy distributions, referred to as the cross-family model $\pi_{\text{cf}}$.
We then use the frozen $\pi_{\text{cf}}$ to construct an evidence reward computation pipeline that operates on the shared evidence but performs independent reasoning (evidence template in Appendix~\ref{subsec: Evidence Template}).
\begin{align}
\alpha_{\text{cf}}&\sim \pi_{\text{cf}}( \cdot \mid q, e ), \notag\\
r^{\text{e}}_{\phi}(q,o) &= \mathrm{F1}\left( \alpha_{\text{cf}},\ \alpha_{\text{gold}} \right)
\end{align}
First, the $\pi_{\text{cf}}$ shares the same evidence $e$ with $\pi_{\theta}$ and generates a Cross-Model Answer $\alpha_{\text{cf}}$ based on $q$.
We then apply the same answer reward computation process to this Cross-Model Answer $\alpha_{\text{cf}}$.
Since $\alpha_{\text{cf}}$ is produced by a different family model than the policy model, it facilitates the mitigation of answer bias introduced by the policy model’s inherent preferences.
As a result, the reward signal more objectively reflects the underlying factual quality of the evidence.

\paragraph{Format reward.}
Format reward ensures that generated content adheres to structural conventions and remains parseable for downstream use~\cite{guo2025deepseek}.
Specifically, we enforce that the evidence appear in exactly one box, marked by \evi{and}.
Similarly, we require the final answer $\alpha$ to be enclosed in exactly one \answer{and} box.
We formalize the calculation pipeline of the format reward $r^{f}_{\phi}$:
\begin{multline}
r^{f}_{\phi}(q,o) = 
(1 - \mathbb{I}_\mathcal{S} )(\gamma_e + \gamma_{\alpha} \cdot \mathbb{I}_\mathcal{A} ) \\+
\mathbb{I}_\mathcal{S} (\gamma_e \cdot \mathbb{I}_\mathcal{E}  + \gamma_{\alpha}\ \cdot \mathbb{I}_\mathcal{A} ),
\end{multline}
where $\mathbb{I}_\mathcal{S}$, $\mathbb{I}_\mathcal{A}$, and $\mathbb{I}_\mathcal{E}$ are indicator functions denoting whether retrieval is triggered, the answer is well-formatted, and the evidence is well-formatted, respectively.
We present the formula for the overall reward $r_{\phi}$. $\gamma_e$ and $\gamma_a$ are reward values.
\begin{align}
r_{\phi} &= r^{\alpha}_{\phi} + r^{e}_{\phi} + r^{f}_{\phi} \\
\text{s.t.}\quad 
&r^{\alpha}_{\phi},\ r^{e}_{\phi} \in [0, 1], \\
&r^{f}_{\phi} \in \{0,\gamma_{e},\gamma_{a},\gamma_e+\gamma_a\}
\end{align}

\subsubsection{Mask and non-mask strategy}
\label{subsubsec: Mask strategy}
\paragraph{Mask strategy for retrieved documents.}
In our rollout sequences, we mix model-generated tokens with externally retrieved documents. Treating the retrieved documents $D_{k}$ as part of the model’s behavior for loss computation can introduce gradient noise and lead to a misalignment with the intended optimization objective.
Therefore, we introduce a loss masking strategy that masks out the search-derived tokens in the observation, ensuring that the optimization objective is applied only to the tokens generated by the policy model during reasoning.
\paragraph{Non-mask strategy for evidence.}
Only a limited number of factual segments can be incorporated into the reasoning chain, resulting in insufficient utilization of external knowledge sources.
To address this, we apply a non-masking strategy to the evidence $e$, aiming to fully utilize external knowledge and enhance the model’s capability for knowledge integration.
Specifically, in the multi-turn reasoning-search interaction, the evidence generated by the model is based on the retrieved information and produced according to the model’s policy distribution, making it eligible for the gradient update.
This strategy allows the evidence to participate in training, guiding the model to more effectively learn how to select, understand, and integrate external knowledge.
It enhances the model’s ability to ground its reasoning on evidence.

\section{Experimental Setup}
\label{Section: Experimental Setup}

\subsection{Datasets \& Metrics, and \& Search Tools}
\label{subsec: Datasets Settings}
\paragraph{Datasets.}
We conduct extensive experiments on seven datasets, covering both complex multi-hop and simpler single-hop QA tasks.
The multi-hop QA serves to evaluate whether R-Search can handle complex logic- and knowledge-intensive questions. The single-hop QA assesses its ability to address knowledge-intensive questions and explore its robustness across questions with varying levels of complexity.
For \textbf{multi-hop} task, we adopt four challenging datasets: \textbf{HotpotQA}~\cite{DBLP:conf/emnlp/Yang0ZBCSM18}, \textbf{2WikiMultiHopQA (2WikiMQA)}~\cite{DBLP:conf/coling/HoNSA20}, \textbf{MuSiQue}~\cite{DBLP:journals/tacl/TrivediBKS22}, and \textbf{Bamboogle}~\cite{DBLP:conf/emnlp/PressZMSSL23}.
For \textbf{single-hop} task, we select three factoid-based QA datasets, including \textbf{NQ}~\cite{DBLP:journals/tacl/KwiatkowskiPRCP19}, \textbf{PopQA}~\cite{DBLP:conf/acl/MallenAZDKH23}, and \textbf{TriviaQA}~\cite{DBLP:conf/acl/JoshiCWZ17}.
The dataset characteristics, versions, and sizes are provided in Appendix~\ref{subsec-app: Datasets Settings}.

\paragraph{Evaluation Metrics.}
Following FLARE~\cite{DBLP:conf/emnlp/JiangXGSLDYCN23}, we adopt two standard evaluation metrics for QA tasks: \textbf{F1-Score} (EM) and \textbf{Exact Match} (EM) for all datasets.
EM is a more stringent metric than F1, as it measures string-level exact matches between the normalized prediction and the golden answer.

\paragraph{Search Tools.}
Effective search actions require appropriate retrieval sources and methods.
We use a dense retriever with the E5 model for all datasets.
For single-hop QA datasets and Bamboogle, we use the 2018 Wikipedia dump as the corpus for retrieving open-domain knowledge.
For the remaining three multi-hop datasets, we use the Wikipedia corpora version released by ~\cite{DBLP:conf/acl/TrivediBKS23}, each aligned with its corresponding dataset.

\subsection{Baselines and Backbone LLMs}
\label{subsec: Baselines Settings}
In our experiments, we conduct comparisons across five types of baselines.
In \textbf{Naive Generation} (NG), we evaluate the ability of LLMs to answer questions using only their internal parametric knowledge.
Vanilla RAG extends NG by adding a one-time search step, forming the simplest retrieval-and-generation pipeline.
For \textbf{Branching RAG}, we use SuRe as the baseline.
Furthermore, we compare with \textbf{Multi-Step RAG} (MSRAG),
including Iter-Retgen~\cite{DBLP:conf/emnlp/ShaoGSHDC23} and IRCoT~\cite{DBLP:conf/acl/TrivediBKS23}, which continuously perform iterative reasoning and search actions to derive the final answer.
We also compare R-Search with the recent advanced RAG method, \textbf{Adaptive RAG} (ARAG),
including FLARE~\cite{DBLP:conf/emnlp/JiangXGSLDYCN23} and Adaptive-RAG~\cite{DBLP:conf/naacl/JeongBCHP24}.
Unlike MSRAG, ARAG leverages the reasoning capabilities of LLMs to actively decide when and what to retrieve, allowing for a more advanced and flexible agentic reasoning–search interaction.
In addition to ARAG, we also compare with Search-R1, a concurrent method that leverages RL to improve the reasoning–search interaction capability of LLMs.
We categorize such methods as \textbf{RAG+RL}.
For Search-R1, we align the training parameters and datasets with those used in our method to ensure a fair comparison. For the other baseline methods, we evaluate their performance using FlashRAG~\cite{DBLP:journals/corr/abs-2405-13576}.
\begin{table*}[ht!]
    \renewcommand{\arraystretch}{0.9}
  \setlength{\tabcolsep}{3pt}
    \fontsize{8}{8.5}\selectfont
  \centering
    \begin{tabular}{c|ccccccccc|ccccccc|c}
    \toprule
    \multirow{3}[6]{*}{\textbf{Method}} & \multicolumn{9}{c|}{\textbf{Multi-Hop QA}}                             & \multicolumn{7}{c|}{\textbf{Single-Hop QA}} \\
\cmidrule(rl){2-10}\cmidrule(rl){11-17}          & \multicolumn{2}{c}{HotpotQA$^{\dagger}$} & \multicolumn{2}{c}{2WikiMQA$^{*}$} & \multicolumn{2}{c}{MuSiQue$^{\dagger}$} & \multicolumn{2}{c}{Bamboogle$^{\dagger}$} &       & \multicolumn{2}{c}{NQ$^{\dagger}$} & \multicolumn{2}{c}{TriviaQA$^{\dagger}$} & \multicolumn{2}{c}{PopQA$^{\dagger}$} &  \\
\cmidrule(rl){2-3}\cmidrule(rl){4-5}\cmidrule(rl){6-7}\cmidrule(rl){8-9}\cmidrule(rl){10-10}\cmidrule(rl){11-12}\cmidrule(rl){13-14}\cmidrule(rl){15-16}\cmidrule(rl){17-17}\cmidrule(rl){18-18}         & EM    & F1    & EM    & F1    & EM    & F1    & EM    & F1    & Avg.  & EM    & F1    & EM    & F1    & EM    & F1    & Avg. & Overall Avg.\\
    \midrule
        \rowcolor[rgb]{ .851,  .851,  .851}\multicolumn{18}{c}{\textbf{Qwen-2.5-3B-Instruct}} \\
    LLM w/o Search & 15.0  & 20.6  & 24.4  & 27.8  & 1.4   & 7.2   & 2.4   & 9.5   & 13.5  & 9.8   & 17.8  & 32.0  & 37.7  & 12.8  & 16.4  & 21.1 & 16.8 \\
    Vanilla RAG & 34.0  & 43.2  & 33.6  & 38.0  & 5.6   & 11.7  & 9.6   & 19.6  & 24.4  & 37.2  & 46.7  & 58.0  & 66.8  & 39.4  & 46.7  & 49.1 &35.0 \\
    SuRe  & 29.2  & 37.8  & 26.8  & 32.4  & 4.2   & 8.9   & 7.2   & 15.6  & 20.3  & 36.4  & 44.8  & 57.0  & 64.4  & 43.2  & 47.2  & 48.8 &32.5 \\
    Iter-Retgen & 34.4  & 43.4  & 33.2  & 38.2  & 8.2   & 14.8  & 12.0  & 20.2  & 25.5  & 38.0  & 47.4  & 60.2  & 68.5  & 43.2  & 49.4  & 51.1 &36.5 \\
    IRCoT & 39.0  & 50.4  & 35.8  & 46.0  & 9.2   & 17.6  & 23.2  & 33.3  & 31.8  & 23.4  & 35.4  & 45.8  & 56.5  & 31.6  & 41.5  & 39.0 &34.9 \\
    FLARE & 14.0  & 20.4  & 24.2  & 27.5  & 1.0   & 6.5   & 3.2   & 9.0   & 13.2  & 9.2   & 16.5  & 32.2  & 38.0  & 12.0  & 15.7  & 20.6 &16.4 \\
    Adaptive-RAG & 38.0  & 49.0  & 35.0  & 43.9  & 25.4 & 35.6 & 24.0  & 32.8  & 35.5 & 37.2  & 46.7  & 55.4  & 64.4  & 35.8  & 43.8  & 47.2 &38.0 \\
    Search-R1 & 46.2  & 57.8  & 58.8  & 68.1   & 24.4   & 32.9   & 41.6   & 53.9   & 48.0   & 34.4  & 44.1   & 56.6   & 63.2   & 37.0  & 43.5   & 46.5 &47.3 \\
    \rowcolor[rgb]{0.85, 0.985, 0.985}\textbf{R-Search (Ours)} & 43.4 	& 54.4  & 65.0 	& 72.6  & 25.8  & 34.8  & 37.6  & 49.8  & 47.9  & 35.2 & 46.0  & 56.0  & 64.0  & 37.0  & 44.9  & 47.2 &47.6
    \\
    \midrule
    \rowcolor[rgb]{ .851,  .851,  .851}\multicolumn{18}{c}{\textbf{Qwen-2.5-7B-Instruct}} \\
    LLM w/o Search & 19.6  & 26.7  & 23.8  & 28.1  & 3.8   & 11.3  & 11.2  & 19.7  & 18.0  & 13.8  & 21.9  & 46.0  & 52.2  & 15.6  & 19.6  & 28.2 &22.4 \\
    Vanilla RAG & 37.4  & 48.1  & 35.4  & 40.5  & 7.2   & 14.6  & 20.8  & 29.7  & 29.2  & 35.0  & 46.7  & 60.0  & 68.5  & 37.6  & 47.6  & 49.2 &37.8\\
    SuRe  & 33.8  & 43.7  & 25.6  & 32.5  & 6.8   & 13.3  & 17.6  & 29.2  & 25.3  & \textbf{42.0} & \textbf{50.8} & 60.0  & 69.1  & \textbf{45.6} & \textbf{50.0} & \textbf{52.9} &37.1 \\
    Iter-Retgen & 42.8  & 53.1  & 37.4  & 43.4  & 10.6  & 19.9  & 22.4  & 31.0  & 32.6  & 37.8  & 48.4  & 61.2  & 69.5  & 38.6  & 46.8  & 50.4 &40.2 \\
    IRCoT & 40.4  & 53.7  & 34.2  & 45.5  & 9.0   & 17.3  & 20.0  & 32.3  & 31.6  & 19.6  & 35.5  & 55.2  & 66.2  & 33.0  & 43.6  & 42.2 &36.1 \\
    FLARE & 17.8  & 24.8  & 22.6  & 27.5  & 3.6   & 11.4  & 12.0  & 19.4  & 17.4  & 13.4  & 21.6  & 42.0  & 48.7  & 16.0  & 19.8  & 26.9 &21.5 \\
    Adaptive-RAG & 42.4  & 55.3  & 33.8  & 42.4  & 9.0   & 16.9  & 20.8  & 32.5  & 31.6  & 35.0  & 46.7  & 58.8  & 67.4  & 35.8  & 45.7  & 48.2 &38.8 \\
    Search-R1 & \underline{48.4}  & \underline{60.9}  & \underline{67.0}  & \underline{75.4}  & \underline{25.8}  & \underline{36.2}  & \textbf{47.2} & \textbf{58.4}  & \underline{52.4}  & \underline{39.8}  & \underline{49.1}  & \textbf{65.0}  & \underline{70.8}  & 41.0  & 46.9   & \underline{52.1}  &\underline{52.3} \\
    \rowcolor[rgb]{0.85, 0.985, 0.985}\textbf{R-Search (Ours)} & \textbf{52.2} & \textbf{64.4} & \textbf{69.8} & \textbf{77.7} & \textbf{31.4} & \textbf{41.6} & \underline{42.4}  & \underline{57.6} & \textbf{54.6} & 38.0  & \underline{49.1}  & \underline{64.2} & \textbf{71.7} & \underline{41.8}  & \underline{48.1}  & \underline{52.1} &\textbf{53.6}\\
    \bottomrule
    \end{tabular}%
    \caption{\textbf{Results (\%) of overall performance.}
    Bold and underlined values represent the highest and second-highest results, respectively. $\dagger$ and $*$ indicate in-domain and out-of-domain datasets.
    }
    \label{Overall Performance}
\end{table*}
% \paragraph{LLMs.}
For backbone LLMs, we train two open-source models via RL, i.e., \textbf{Qwen-2.5-3B/7B-Instruct}~\cite{yang2024qwen2}. To construct the evidence reward, we use a different family open-source model, Llama-3.2-3B-Instruct~\cite{Llama-3.2}, which follows a different policy distribution, to generate cross-model answers based on the shared evidence.
\subsection{Implementation Details}
\label{subsec: Implementation Details}
During evaluation, we align the top-$k$ retrieval setting to 5 across all methods to ensure fair comparison. We also employ vLLM~\cite{DBLP:conf/sosp/KwonLZ0ZY0ZS23} to accelerate inference.
For GRPO and PPO training, we use only the 2WikiMQA training set and train on 8×A100 80GB GPUs. Primary training hyperparameters include a maximum total of 195 training steps and a batch size of 256.
Following Search-R1, we set the retrieval top-$k$ during training to 3, and configure the learning rate and warm-up ratio to 1e-6 and 0.95, respectively.
During the rollout process, we sample 5 responses for each input prompt, with the KL divergence coefficient $\beta$ set to 0.001, and fix both $\gamma_{e}$ and $\gamma_{\alpha}$ to 0.2.
\section{Results and Analysis}
\label{Section: Results and Analysis}
\subsection{Overall Performance}
In Table~\ref{Overall Performance}, we present the overall performance of various baselines and R-Search (case study in Appendix~\ref{sec: Case Study}).

\noindent\textbf{R-Search facilitates deep knowledge exploration.}
Compared to Vanilla RAG and LLM w/o search, our framework achieves up to a 37.2\% improvement on complex multi-hop QA tasks. It also delivers gains of up to 4.2\% on simpler single-hop datasets that require less reasoning.
These results demonstrate that our framework effectively supports deep knowledge exploration and ensures a stable reasoning–search interaction process.

\noindent\textbf{R-Search generalizes well to questions with both simple and complex reasoning–search requirements.}
We observe that branching RAG (e.g., SuRe) performs competitively on simple single-hop QA tasks. However, its performance drops sharply on multi-hop questions (e.g., MuSiQue) with higher search and reasoning demands.
In contrast, our method excels at handling questions with long reasoning chains and complex retrieval needs, achieving up to a 45.2\% improvement over the branching RAG. These results suggest that R-Search not only achieves strong performance but also adapts well to questions of varying complexity.

\noindent\textbf{R-Search achieves more stable logical reasoning and more targeted, in-depth retrieval than advanced RAG variants.}
We compare our method with the multi-step and ARAG methods.
While these multi-turn RAG methods help mitigate the knowledge limitations of Vanilla RAG through multiple searches, we observe performance instability on more complex multi-hop datasets, such as MuSiQue.
This is partly because multi-step RAG continuously interleaves search and reasoning, which can introduce irrelevant passages. Although adaptive RAG allows the LLM to decide when to retrieve, it often suffers from mismatches between the model’s internal knowledge and actual retrieval needs. In contrast, our method supports more effective and stable reasoning processes, along with deeper search, making it better suited for handling complex reasoning–search interactions.

\noindent\textbf{R-Search effectively optimizes the interaction between search and reasoning through multi-reward signals, enabling stronger performance gains on highly complex tasks.}
We also evaluate against a concurrent RAG+RL approach, Search-R1. Results show that R-Search consistently outperforms Search-R1 in most cases. Notably, R-Search surpasses Search-R1 by 5.6\% on the highly challenging MuSiQue dataset.

\noindent\textbf{Impact analysis of using different backbone LLMs.}
R-Search with the larger model generally outperforms it with the smaller model, mainly due to the stronger instruction-following ability of the larger model and its richer internal knowledge.
Plus, R-Search with the smaller model still outperforms other RAG methods in most cases, especially on multi-hop QA datasets.
R-Search also delivers competitive performance against Search-R1.

\begin{figure}
  \centering
  \includegraphics[width=0.4\textwidth]{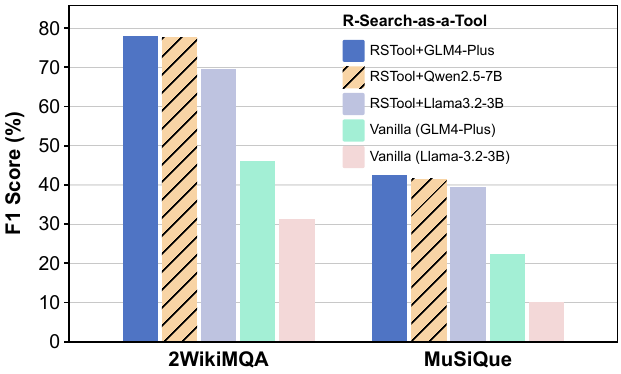}
  \caption{\textbf{R-Search-as-a-Tool.}}
  \label{fig:R-Search-as-a-Tool}
\end{figure}

\noindent\textbf{R-Search-as-a-Tool.}
The shareable evidence in R-Search can also serve as a pluggable component, easily transferred to other models for downstream answer generation. We refer to this transferable functionality as R-Search-as-a-Tool (\textbf{RSTool}).
As shown in Figure~\ref{fig:R-Search-as-a-Tool}, we evaluate the effectiveness of applying the shared evidence to different downstream generation models on two complex multi-hop datasets. We test on both a powerful black-box model, GLM-4-Plus~\cite{DBLP:journals/corr/abs-2406-12793}, and an open-source model from a different family than the policy model, Llama-3.2-3B. For example, "RSTool + GLM-4-Plus" indicates that GLM-4-Plus generates answers using the shareable evidence generated by the trained Qwen-2.5-7B-Instruct model.
The results show that applying the shareable evidence to downstream models with different sizes and architectures consistently improves performance, achieving significant gains over Vanilla. This suggests the shared evidence captures high-quality, comprehensive knowledge effectively distilled from the prior reasoning–retrieval interaction. In practical scenarios, RSTool allows users to offload the high-cost reasoning–search process, often constrained by API token limits, to local models, substantially reducing potential overhead.
\subsection{Ablation Study}
In Table~\ref{Ablation Study} (more results in Appendix~\ref{subsec-app: More Results}), we conduct ablation studies to further analyze the effectiveness of the evidence mechanism.
"R-Search w/o Evidence" refers to a variant where the evidence is removed from rollout, and the evidence-related rewards in both the evidence and format reward are disabled. We observe a clear performance drop without the evidence component and its associated rewards, confirming its importance. This drop is especially pronounced on complex multi-hop datasets compared to single-hop datasets.
On the one hand, during the reasoning stage, multi-step search and long reasoning chains are prone to introducing irrelevant information.
Evidence helps the LLM identify and extract the most relevant knowledge from a global perspective, preventing it from over-focusing on a single reasoning path and missing other valuable clues.
\begin{figure*}[t]
  \centering
\includegraphics[width=\textwidth]{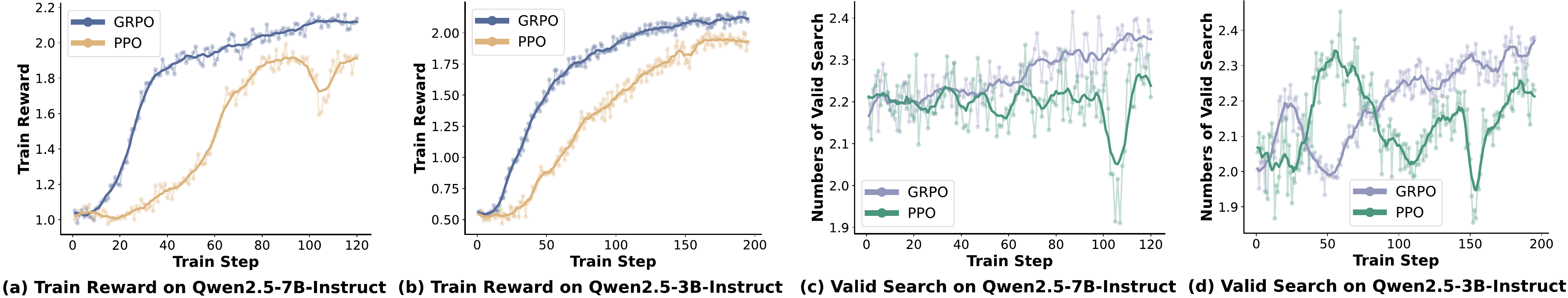}
  \caption{\textbf{Analysis of training reward and number of valid searches on Qwen-2.5-7B/3B-Instruct models.}
  The semi-transparent and the solid lines indicate raw samples and the smoothed trend.}
  \label{fig:Train}
\end{figure*}
\begin{table}[ht]
  \centering
      \renewcommand{\arraystretch}{0.9}
    \setlength{\tabcolsep}{2pt}
    \fontsize{7}{8}\selectfont
    \begin{tabular}{cccccccc}
    \toprule
    \multirow{2}[2]{*}{\textbf{Method}} & \multicolumn{4}{c}{\textbf{Multi-Hop QA}} & \multicolumn{3}{c}{\textbf{Single-Hop QA}} \\
\cmidrule(lr){2-5}\cmidrule(lr){6-8}           & Hot.  & 2Wiki & MuSi. & Bamb. & NQ    & Tri.  & Pop. \\
    \midrule
    \multicolumn{8}{c}{\textbf{Qwen2.5-3B-Instruct}} \\
    R-Search w/o Evidence & \cellcolor[rgb]{.953,  .886,  1}53.4 & \cellcolor[rgb]{.953,  .886,  1}66.6 & \cellcolor[rgb]{.953,  .886,  1}33.3 & \cellcolor[rgb]{.953,  .886,  1}42.3 & \cellcolor[rgb]{.953,  .886,  1}44.6 & \cellcolor[rgb]{.953,  .886,  1}62.3 & \cellcolor[rgb]{.953,  .886,  1}41.6 \\
    R-Search & \cellcolor[rgb]{.882,  .984,  .992}54.4 & \cellcolor[rgb]{.882,  .984,  .992}72.6 & \cellcolor[rgb]{.882,  .984,  .992}34.8 & \cellcolor[rgb]{.882,  .984,  .992}49.8 & \cellcolor[rgb]{.882,  .984,  .992}46.0 & \cellcolor[rgb]{.882,  .984,  .992}64.0 & \cellcolor[rgb]{.882,  .984,  .992}44.9 \\
    \midrule
    \multicolumn{8}{c}{\textbf{Qwen2.5-7B-Instruct}} \\
    R-Search w/o Evidence & \cellcolor[rgb]{.953,  .886,  1}61.9 & \cellcolor[rgb]{.953,  .886,  1}77.5 & \cellcolor[rgb]{.953,  .886,  1}39.6 & \cellcolor[rgb]{.953,  .886,  1}55.9 & \cellcolor[rgb]{.953,  .886,  1}47.3 & \cellcolor[rgb]{.953,  .886,  1}70.2 & \cellcolor[rgb]{.953,  .886,  1}48.0 \\
    R-Search & \cellcolor[rgb]{.882,  .984,  .992}64.4 & \cellcolor[rgb]{.882,  .984,  .992}77.7 & \cellcolor[rgb]{.882,  .984,  .992}41.6 & \cellcolor[rgb]{.882,  .984,  .992}57.6 & \cellcolor[rgb]{.882,  .984,  .992}49.1 & \cellcolor[rgb]{.882,  .984,  .992}71.7 & \cellcolor[rgb]{.882,  .984,  .992}48.1 \\
    \bottomrule
    \end{tabular}%
    \caption{\textbf{Results (\%) of ablation study.}
     "Hot." stands for HotpotQA; other dataset names are similarly abbreviated.
    \colorbox[rgb]{.882,  .984,  .992}{"Blue"}, and \colorbox[rgb]{.953,  .886,  1}{"purple"} are the highest and lowest values.
  }
  \label{Ablation Study}
\end{table}%
\begin{table}[ht]
  \centering
      \renewcommand{\arraystretch}{0.9}
  \setlength{\tabcolsep}{3pt}
    \fontsize{7}{8}\selectfont
    \begin{tabular}{cccccccc}
    \toprule
    \multirow{2}[2]{*}{\textbf{Method}} & \multicolumn{4}{c}{\textbf{Multi-Hop QA}} & \multicolumn{3}{c}{\textbf{Single-Hop QA}} \\
\cmidrule(lr){2-5}\cmidrule(lr){6-8}          & Hot.  & 2Wiki & MuSi. & Bamb. & NQ    & Tri.  & Pop. \\
    \midrule
    \multicolumn{8}{c}{\textbf{Qwen2.5-3B-Instruct}} \\
    R-Search (PPO) & \cellcolor[rgb]{.953,  .886,  1}52.4 & \cellcolor[rgb]{.953,  .886,  1}60.3 & \cellcolor[rgb]{.953,  .886,  1}33.5 & \cellcolor[rgb]{.953,  .886,  1}50.0 & \cellcolor[rgb]{.953,  .886,  1}43.7 & \cellcolor[rgb]{.953,  .886,  1}63.1 & \cellcolor[rgb]{.953,  .886,  1}44.2 \\
    R-Search (GRPO) & \cellcolor[rgb]{.882,  .984,  .992}54.4 & \cellcolor[rgb]{.882,  .984,  .992}72.6 & \cellcolor[rgb]{.882,  .984,  .992}34.8 & \cellcolor[rgb]{.882,  .984,  .992}49.8 & \cellcolor[rgb]{.882,  .984,  .992}46.0 & \cellcolor[rgb]{.882,  .984,  .992}64.0 & \cellcolor[rgb]{.882,  .984,  .992}44.9 \\
    \midrule
    \multicolumn{8}{c}{\textbf{Qwen2.5-7B-Instruct}} \\
    R-Search (PPO) & \cellcolor[rgb]{.953,  .886,  1}58.7 & \cellcolor[rgb]{.953,  .886,  1}68.9 & \cellcolor[rgb]{.953,  .886,  1}37.3 & \cellcolor[rgb]{.953,  .886,  1}53.7 & \cellcolor[rgb]{.953,  .886,  1}47.3 & \cellcolor[rgb]{.953,  .886,  1}70.4 & \cellcolor[rgb]{.953,  .886,  1}45.5 \\
    R-Search (GRPO) & \cellcolor[rgb]{.882,  .984,  .992}64.4 & \cellcolor[rgb]{.882,  .984,  .992}77.7 & \cellcolor[rgb]{.882,  .984,  .992}41.6 & \cellcolor[rgb]{.882,  .984,  .992}57.6 & \cellcolor[rgb]{.882,  .984,  .992}49.1 & \cellcolor[rgb]{.882,  .984,  .992}71.7 & \cellcolor[rgb]{.882,  .984,  .992}48.1 \\
    \bottomrule
    \end{tabular}%
    \caption{\textbf{PPO vs. GRPO Performance (\%).}}
  \label{PPO vs. GRPO}
\end{table}%
On the other hand, evidence provides intermediate reward signals along the long reasoning–search trajectory, guiding the model to prioritize the reliability and completeness of intermediate factual content, rather than relying on speculative strategies that may yield unintended correct answers.

\subsection{Analysis}
\paragraph{Performance and reward.}
We conduct experiments using both PPO and GRPO on Qwen-2.5-3B-Instruct and Qwen-2.5-7B-Instruct models.
Table~\ref{PPO vs. GRPO} presents the overall performance results, while Figure~\ref{fig:Train} illustrates the training dynamics and trends of key statistics during optimization.
In terms of performance, as shown in Table~\ref{PPO vs. GRPO} (more results in Appendix~\ref{subsec-app: More Results}), both GRPO and PPO lead to strong results. These findings indicate that both algorithms are suitable for optimizing R-Search and demonstrate the general applicability of our framework.
Furthermore, the results show that GRPO generally outperforms PPO, particularly on larger models and more complex multi-hop tasks.
A potential reason is that reward signals tend to be sparse and delayed, making PPO more prone to getting stuck in local optima in complex tasks.
In addition, larger models with stronger instruction-following capabilities can execute more effective reasoning–retrieval trajectories, leading to better performance. During training, as shown in Figures ~\ref{fig:Train}(a) and ~\ref{fig:Train}(b), we observe that GRPO converges faster than PPO and reaches a higher reward ceiling.
This is because PPO relies on an actor-critic architecture, where the critic is unstable and requires a warm-up phase in the early stage, easily introducing noise, while GRPO bypasses the limitations of value function estimation, making it more likely to achieve a higher reward ceiling.
Therefore, we recommend using larger LLMs together with GRPO when applying R-Search, as this setup is more likely to result in faster convergence and better final performance.
\paragraph{Number of valid searches.}
Figures ~\ref{fig:Train}(c) and ~\ref{fig:Train}(d) show the learning dynamics and trends in the number of valid searches as training progresses. We observe that the trained models tend to trigger more retrieval steps, engaging in more rounds of reasoning–search interaction and enabling deeper exploration of external knowledge.

\section{Conclusion}
We propose R-Search, a novel RL-based RAG framework that autonomously optimizes reasoning-search trajectories via multi-reward signals, seamlessly integrating reasoning and search for complex problem solving. Extensive experiments on seven benchmarks demonstrate the superiority of R-Search. Additionally, R-Search generates explicit evidence when invoking external search, enabling it to function as a modular search tool. This design enhances the deep reasoning-search interaction and allows efficient offloading of resource-intensive search processes to local models.

\clearpage
\section*{Limitations}
To demonstrate the strong generalization capability of R-Search, we only use the 2WikiMQA training dataset during the training phase. Although our method has achieved significant performance improvements on both in-domain and out-of-domain tasks under this setup, we acknowledge that incorporating more high-quality knowledge from diverse domains during training may further enhance the model's performance. Therefore, future work should explore integrating more diverse and high-quality knowledge sources to further improve the effectiveness of R-Search.

\bibliography{custom}

\clearpage

\appendix

\section*{Appendix}
% \label{sec-app:appendix}
\section{Experiment and Result}
\label{sec-app: Experiment and result}

\subsection{Datasets}
\label{subsec-app: Datasets Settings}
We conduct extensive experiments on seven datasets, covering both complex multi-hop and simpler single-hop QA tasks.
The multi-hop QA task serves to evaluate whether R-Search can handle complex logic- and knowledge-intensive questions. The single-hop QA task assesses its ability to address knowledge-intensive questions and explore its robustness across questions with varying levels of complexity.
For \textbf{Multi-hop} task, we adopt four challenging multi-hop QA datasets: \textbf{HotpotQA}~\cite{DBLP:conf/emnlp/Yang0ZBCSM18}, \textbf{2WikiMultiHopQA (2WikiMQA)}~\cite{DBLP:conf/coling/HoNSA20}, \textbf{MuSiQue}~\cite{DBLP:journals/tacl/TrivediBKS22}, and \textbf{Bamboogle}~\cite{DBLP:conf/emnlp/PressZMSSL23}.
The datasets require models to start from the original question and explore a logical, knowledge-driven multi-hop reasoning path to answer each intermediate sub-question and reach the final answer.
We follow the train and test splits released by~\cite{DBLP:conf/acl/TrivediBKS23} for the first three datasets, each with 500 test samples. Bamboogle evaluation uses all 125 test samples provided by FlashRAG~\cite{DBLP:journals/corr/abs-2405-13576}.
For \textbf{Single-hop} task, we select three factoid-based QA datasets, including \textbf{NQ}~\cite{DBLP:journals/tacl/KwiatkowskiPRCP19}, \textbf{PopQA}~\cite{DBLP:conf/acl/MallenAZDKH23}, and \textbf{TriviaQA}~\cite{DBLP:conf/acl/JoshiCWZ17}. These datasets require models to collect specific passages and identify key factual information to answer the questions correctly. We use the test sets provided by FlashRAG and randomly sample 500 examples from each dataset for evaluation.

\subsection{Results}
\label{subsec-app: More Results}
We present the results of all evaluation metrics for the ablation study and the comparison between PPO and GRPO in Table~\ref{app: Ablation Study} and Table~\ref{app: PPO vs. GRPO}.
The trends of EM are consistent with those of F1.

\subsection{More Implementation Details}
\label{subsec-app: Implementation Details}
During inference, we set the temperature to 0.1 across all models to reduce uncertainty.

\section{Template}
\label{sec:Template}

\subsection{System Template}
\label{subsec: System Template}
We design a system template for the rollout phase to guide the model through the complete interaction process—from receiving the input question $q$ to generating the final answer $\alpha$.
This template covers four key stages: reasoning, retrieval, evidence integration, and answer generation.
Notably, for questions that do not require external knowledge, the LLM automatically determines that search is unnecessary.
In such cases, the template instructs the model to perform only reasoning and answer generation.
During training, we initialize the training process with the system template and the user's question.
It is worth noting that we do not impose a manual separation between reasoning and retrieval (e.g., by encapsulating reasoning within a \think{and} tag); Instead, we provide a high-level instruction that guides the model to explain its thought process before each action, without imposing any specific format for reasoning text.
This stems from the fact that LLM generation is inherently a form of reasoning, where every generated token can be viewed as part of a thought chain for question-solving.
For parts such as \hsearch{}, \hobs{}, \hevi{}, and \hanswer, which are distinct from general thought reasoning, we introduce specific token tags to mark them, allowing the LLM to recognize the boundaries between different functional segments.

\subsection{Evidence Template}
\label{subsec: Evidence Template}
Table~\ref{app: Evidence template} presents the instruction template used for generating evidence during rollout. Before outputting the final answer $\alpha$, we instruct the LLM to generate evidence based on the original question $q$ and all previously retrieved texts (provided in the \obs{and}).
This evidence helps the LLM rethink the retrieved information from a global perspective and focus on key factual knowledge.
By seamlessly integrating the evidence into the reasoning process, we facilitate a deeper interaction between reasoning and retrieved knowledge.

\section{Case Study}
\label{sec: Case Study}
Tables~\ref{tab-app: case ppo-musique-2hop} and~\ref{tab-app: case grpo-2wiki-4hop} show examples of the reasoning process from RL-trained models on 2-hop and more complex 4-hop questions. In these cases, R-Search uses multi-stage, multi-type rewards to improve the reasoning–retrieval process.
The model usually starts by generating a general reasoning plan, retrieves information when needed, and uses intermediate conclusions to guide the next retrieval.
The evidence generated through this interaction is clear, well-structured, and informative, making it easy to transfer to downstream models for final answer generation.

\clearpage

\begin{table*}[ht]
\centering
\begin{tabular}{p{13cm}}
\toprule
\toprule
Answer the question based on the given passages.
Only give me the answer and do not output any other words.
\\
% \midrule
The following are given passages: \{\texttt{evidence}\}
\\
Question: \{\texttt{query}\}
\\
Answer:
\\
\bottomrule
\bottomrule
\end{tabular}
\caption{\textbf{Evidence template.}}
\label{app: Evidence template}
\end{table*}

\begin{table*}[ht!]
  \setlength{\tabcolsep}{4pt}
    \fontsize{8}{9}\selectfont
  \centering
    \begin{tabular}{ccccccccccccccc}
    \toprule
    \toprule
    \multirow{3}[6]{*}{\textbf{Method}} & \multicolumn{8}{c}{\textbf{Multi-Hop QA}}                     & \multicolumn{6}{c}{\textbf{Single-Hop QA}} \\
\cmidrule(rl){2-9}\cmidrule(rl){10-15}         & \multicolumn{2}{c}{HotpotQA} & \multicolumn{2}{c}{2WikiMQA} & \multicolumn{2}{c}{MuSiQue} & \multicolumn{2}{c}{Bamboogle} & \multicolumn{2}{c}{NQ} & \multicolumn{2}{c}{TriviaQA} & \multicolumn{2}{c}{PopQA} \\
\cmidrule(rl){2-3}\cmidrule(rl){4-5}\cmidrule(rl){6-7} \cmidrule(rl){8-9} \cmidrule(rl){10-11} \cmidrule(rl){12-13} \cmidrule(rl){14-15}           & EM    & F1    & EM    & F1    & EM    & F1    & EM    & F1    & EM    & F1    & EM    & F1    & EM    & F1 \\
    \midrule
    \multicolumn{15}{c}{\textbf{Qwen-2.5-3B-Instruct}} \\
    R-Search w/o Evidence & \cellcolor[rgb]{ .953,  .886,  1}41.0  & \cellcolor[rgb]{ .953,  .886,  1}53.4  & \cellcolor[rgb]{ .953,  .886,  1}58.2  & \cellcolor[rgb]{ .953,  .886,  1}66.6  & \cellcolor[rgb]{ .953,  .886,  1}23.6  & \cellcolor[rgb]{ .953,  .886,  1}33.3  & \cellcolor[rgb]{ .953,  .886,  1}32.0 & \cellcolor[rgb]{ .953,  .886,  1}42.3  & \cellcolor[rgb]{ .953,  .886,  1}34.6  & \cellcolor[rgb]{ .953,  .886,  1}44.6  & \cellcolor[rgb]{ .953,  .886,  1}54.4  & \cellcolor[rgb]{ .953,  .886,  1}62.3  & \cellcolor[rgb]{ .953,  .886,  1}36.0  & \cellcolor[rgb]{ .953,  .886,  1}41.6  \\
    R-Search & \cellcolor[rgb]{ .882,  .984,  .992}43.4 & \cellcolor[rgb]{ .882,  .984,  .992}54.4 & \cellcolor[rgb]{ .882,  .984,  .992}65.0 & \cellcolor[rgb]{ .882,  .984,  .992}72.6  & \cellcolor[rgb]{ .882,  .984,  .992}25.8 & \cellcolor[rgb]{ .882,  .984,  .992}34.8 & \cellcolor[rgb]{ .882,  .984,  .992}37.6 & \cellcolor[rgb]{ .882,  .984,  .992}49.8 & \cellcolor[rgb]{ .882,  .984,  .992}35.2  & \cellcolor[rgb]{ .882,  .984,  .992}46.0 & \cellcolor[rgb]{ .882,  .984,  .992}56.0 & \cellcolor[rgb]{ .882,  .984,  .992}64.0 & \cellcolor[rgb]{ .882,  .984,  .992}37.0 & \cellcolor[rgb]{ .882,  .984,  .992}44.9 \\
    \midrule
    \multicolumn{15}{c}{\textbf{Qwen-2.5-7B-Instruct}} \\
    R-Search w/o Evidence & \cellcolor[rgb]{ .953,  .886,  1}49.4  & \cellcolor[rgb]{ .953,  .886,  1}61.9  & \cellcolor[rgb]{ .953,  .886,  1}69.0  & \cellcolor[rgb]{ .953,  .886,  1}77.5  & \cellcolor[rgb]{ .953,  .886,  1}29.4  & \cellcolor[rgb]{ .953,  .886,  1}39.6 & \cellcolor[rgb]{ .953,  .886,  1}44.0   & \cellcolor[rgb]{ .953,  .886,  1}55.9  & \cellcolor[rgb]{ .953,  .886,  1}36.8  & \cellcolor[rgb]{ .953,  .886,  1}47.3  & \cellcolor[rgb]{ .953,  .886,  1}63.6  & \cellcolor[rgb]{ .953,  .886,  1}70.2  & \cellcolor[rgb]{ .953,  .886,  1}41.4  & \cellcolor[rgb]{ .953,  .886,  1}48.0  \\
    R-Search & \cellcolor[rgb]{ .882,  .984,  .992}52.2  & \cellcolor[rgb]{ .882,  .984,  .992}64.4  & \cellcolor[rgb]{ .882,  .984,  .992}69.8  & \cellcolor[rgb]{ .882,  .984,  .992}77.7  & \cellcolor[rgb]{ .882,  .984,  .992}31.4  & \cellcolor[rgb]{ .882,  .984,  .992}41.6  & \cellcolor[rgb]{ .882,  .984,  .992}42.4   & \cellcolor[rgb]{ .882,  .984,  .992}57.6  & \cellcolor[rgb]{ .882,  .984,  .992}38.0  & \cellcolor[rgb]{ .882,  .984,  .992}49.1  & \cellcolor[rgb]{ .882,  .984,  .992}64.2  & \cellcolor[rgb]{ .882,  .984,  .992}71.7  & \cellcolor[rgb]{ .882,  .984,  .992}41.8  & \cellcolor[rgb]{ .882,  .984,  .992}48.1  \\
    \bottomrule
    \bottomrule
    \end{tabular}%
    \caption{\textbf{Results (\%) of ablation study.}
  \colorbox[rgb]{.882,  .984,  .992}{"Blue"}, and \colorbox[rgb]{.953,  .886,  1}{"purple"} are the highest and lowest values.
  }
  \label{app: Ablation Study}
\end{table*}%

\begin{table*}[ht!]
  \setlength{\tabcolsep}{4pt}
    \fontsize{8}{9}\selectfont
  \centering
    \begin{tabular}{ccccccccccccccccc}
    \toprule
        \toprule
    \multirow{3}[6]{*}{\textbf{Method}} & \multicolumn{9}{c}{\textbf{Multi-Hop QA}}                             & \multicolumn{7}{c}{\textbf{Single-Hop QA}} \\
\cmidrule(rl){2-10}\cmidrule(rl){11-17}          & \multicolumn{2}{c}{HotpotQA} & \multicolumn{2}{c}{2WikiMQA} & \multicolumn{2}{c}{MuSiQue} & \multicolumn{2}{c}{Bamboogle} &       & \multicolumn{2}{c}{NQ} & \multicolumn{2}{c}{TriviaQA} & \multicolumn{2}{c}{PopQA} &  \\
\cmidrule(rl){2-3}\cmidrule(rl){4-5}\cmidrule(rl){6-7}\cmidrule(rl){8-9}\cmidrule(rl){11-12}\cmidrule(rl){13-14}\cmidrule(rl){15-16}          & EM    & F1    & EM    & F1    & EM    & F1    & EM    & F1    & Avg.  & EM    & F1    & EM    & F1    & EM    & F1    & Avg. \\
    \midrule
    \multicolumn{17}{c}{\textbf{Qwen-2.5-3B-Instruct}} \\
    R-Search (PPO) & \cellcolor[rgb]{ .953,  .886,  1}42.0 & \cellcolor[rgb]{ .953,  .886,  1}52.42 & \cellcolor[rgb]{ .953,  .886,  1}50.8 & \cellcolor[rgb]{ .953,  .886,  1}60.3  & \cellcolor[rgb]{ .953,  .886,  1}24.0 & \cellcolor[rgb]{ .953,  .886,  1}33.5 & \cellcolor[rgb]{ .953,  .886,  1}37.6 & \cellcolor[rgb]{ .953,  .886,  1}50.0  & \cellcolor[rgb]{ .953,  .886,  1}43.8  & \cellcolor[rgb]{ .953,  .886,  1}34.0 & \cellcolor[rgb]{ .953,  .886,  1}43.7  & \cellcolor[rgb]{ .953,  .886,  1}54.6  & \cellcolor[rgb]{ .953,  .886,  1}63.1  & \cellcolor[rgb]{ .953,  .886,  1}36.8  & \cellcolor[rgb]{ .953,  .886,  1}44.2  & \cellcolor[rgb]{ .953,  .886,  1}46.1  \\
    R-Search (GRPO) & \cellcolor[rgb]{ .882,  .984,  .992}43.4 & \cellcolor[rgb]{ .882,  .984,  .992}54.4 & \cellcolor[rgb]{ .882,  .984,  .992}65.0 & \cellcolor[rgb]{ .882,  .984,  .992}72.6 & \cellcolor[rgb]{ .882,  .984,  .992}25.8 & \cellcolor[rgb]{ .882,  .984,  .992}34.8 & \cellcolor[rgb]{ .882,  .984,  .992}37.6 & \cellcolor[rgb]{ .882,  .984,  .992}49.8  & \cellcolor[rgb]{ .882,  .984,  .992}47.9 & \cellcolor[rgb]{ .882,  .984,  .992}35.2 & \cellcolor[rgb]{ .882,  .984,  .992}46.0  & \cellcolor[rgb]{ .882,  .984,  .992}56.0 & \cellcolor[rgb]{ .882,  .984,  .992}64.0 & \cellcolor[rgb]{ .882,  .984,  .992}37.0 & \cellcolor[rgb]{ .882,  .984,  .992}44.9 & \cellcolor[rgb]{ .882,  .984,  .992}47.2 \\
    \midrule
    \multicolumn{17}{c}{\textbf{Qwen-2.5-7B-Instruct}} \\
    R-Search (PPO) & \cellcolor[rgb]{ .953,  .886,  1}47.0 & \cellcolor[rgb]{ .953,  .886,  1}58.7  & \cellcolor[rgb]{ .953,  .886,  1}59.8 & \cellcolor[rgb]{ .953,  .886,  1}68.9 & \cellcolor[rgb]{ .953,  .886,  1}27.0 & \cellcolor[rgb]{ .953,  .886,  1}37.3 & \cellcolor[rgb]{ .953,  .886,  1}40.0 & \cellcolor[rgb]{ .953,  .886,  1}53.7 & \cellcolor[rgb]{ .953,  .886,  1}49.1  & \cellcolor[rgb]{ .953,  .886,  1}37.0 & \cellcolor[rgb]{ .953,  .886,  1}47.3 & \cellcolor[rgb]{ .953,  .886,  1}63.2  & \cellcolor[rgb]{ .953,  .886,  1}70.4  & \cellcolor[rgb]{ .953,  .886,  1}37.8 & \cellcolor[rgb]{ .953,  .886,  1}45.5 & \cellcolor[rgb]{ .953,  .886,  1}50.2  \\
    R-Search (GRPO) & \cellcolor[rgb]{ .882,  .984,  .992}52.2  & \cellcolor[rgb]{ .882,  .984,  .992}64.4  & \cellcolor[rgb]{ .882,  .984,  .992}69.8  & \cellcolor[rgb]{ .882,  .984,  .992}77.7  & \cellcolor[rgb]{ .882,  .984,  .992}31.4  & \cellcolor[rgb]{ .882,  .984,  .992}41.6  & \cellcolor[rgb]{ .882,  .984,  .992}42.4   & \cellcolor[rgb]{ .882,  .984,  .992}57.6  & \cellcolor[rgb]{ .882,  .984,  .992}54.6   
    & \cellcolor[rgb]{ .882,  .984,  .992}38.0  & \cellcolor[rgb]{ .882,  .984,  .992}49.1  & \cellcolor[rgb]{ .882,  .984,  .992}64.2  & \cellcolor[rgb]{ .882,  .984,  .992}71.7  & \cellcolor[rgb]{ .882,  .984,  .992}41.8  & \cellcolor[rgb]{ .882,  .984,  .992}48.1 & \cellcolor[rgb]{ .882,  .984,  .992}52.1   \\
    \bottomrule
        \bottomrule
    \end{tabular}%
    \caption{\textbf{PPO vs. GRPO: Performance (\%) comparison.}}
  \label{app: PPO vs. GRPO}
\end{table*}%

\clearpage

\begin{table*}[ht!]
\renewcommand{\arraystretch}{1.2}
\setlength{\tabcolsep}{2pt}
    \fontsize{9}{10}\selectfont
    \centering
    \begin{tabular}{p{15cm}}
\toprule
\toprule
{\bf Question:} When was countrywide bought by the company that bought FleetBoston Financial?
\\\bottomrule
\rowcolor[rgb]{ .900,  .900,  .900}{\textit{"Thought":"}
To answer this question, we need to follow these steps:}
\\
\rowcolor[rgb]{ .900,  .900,  .900}{1.  Identify the company that bought FleetBoston Financial.}
\\
\rowcolor[rgb]{ .900,  .900,  .900}{2.  Determine when this company bought Countrywide.}
\\
\rowcolor[rgb]{ .900,  .900,  .900}{Let's start with step 1.}
\\
\search{FleetBoston Financial was bought by whom?}
\\
\hobs \,\, (Title: "Bank of America") \hlrel{In 2004, Bank of America announced it would purchase Boston - based bank FleetBoston Financial} for \$47 billion in cash and stock.  By merging with Bank of America, all of its banks and branches were given the Bank of America logo.  At the time of merger, FleetBoston was the seventh largest bank in United States with \$197 billion in assets, over 20 million customers and revenue of \$12 billion.
\\
(Title: "Boston") Boston is an intellectual, technological, and political center but has lost some important regional institutions $\dots$
\\
(Title: "Boston") Other important industries are financial services $\dots$
\\
(Title: "The Ritz-Carlton Hotel Company") Blackstone Real Estate Acquisitions of New York bought The Ritz $\dots$
\\
(Title: "RBC Bank") On June 19, 2011, PNC Financial Services agreed to buy RBC Bank USA from Royal Bank of Canada for \$3.45 billion $\dots$
\hobs
\\
\rowcolor[rgb]{ .900,  .900,  .900}{From the observations, we can conclude that FleetBoston Financial was bought by Bank of America in 2004.}
\\
\rowcolor[rgb]{ .900,  .900,  .900}{Now, let's move to step 2.}
\\
\search{When did Bank of America buy Countrywide?}
\\
\hobs \,\,
(Title: "Bank of America Home Loans") On January 11, 2008, Bank of America announced that it planned to purchase Countrywide Financial for \$4.1 billion in stock.  On June 5, 2008, Bank of America Corporation announced it had received approval from the Board of Governors of the Federal Reserve System to purchase Countrywide Financial Corporation.  Then, on June 25, 2008, Countrywide announced it had received the approval of 69\% of its shareholders to the planned merger with Bank of America.  \hlrel{On July 1, 2008, Bank of America Corporation completed its purchase of Countrywide Financial Corporation.}  In 1997, Countrywide had spun off Countrywide Mortgage Investment as an independent company called IndyMac Bank.  Federal regulators seized IndyMac on July 11, 2008, after a week - long bank run $\dots$
\hobs
\\
(Title: "Bank of America Home Loans") Bank of America Home Loans is the mortgage unit of Bank of America$\dots$
\\
(Title: "Bank of America") In 2004, Bank of America announced it would purchase Boston - based bank FleetBoston Financial for \$47 billion in cash and stock$\dots$
\\
(Title: "Bank of America") Its acquisition of investment bank Merrill Lynch in 2008 solidified it as a ``Bulge Bracket bank ''$\dots$
\\
(Title: "General Tire") GenCorp sold General Tire to German tire maker Continental AG in 1987 $\dots$
\\
\rowcolor[rgb]{ .900,  .900,  .900}{From the observations, we can conclude that Bank of America bought Countrywide on July 1, 2008.}
\\
\hevi- FleetBoston Financial was bought by Bank of America in 2004.
\\
- Bank of America bought Countrywide on July 1, 2008.
\hevi
\\
\answer{July 1, 2008} \,\, "\textcolor{green}{\ding{51}}"
\\\bottomrule
\bottomrule
 \end{tabular}
    \caption{{\bf An 2-hop example of R-Search on MuSiQue dataset using PPO.
}
\hlrel{Orange segments} denote those used to support correct answers.
% \hlred{Red segments} represent key incorrect information and erroneous responses collected.
The "\textcolor{green}{\ding{51}}" indicates a correct answer.
}
    \label{tab-app: case ppo-musique-2hop}
\end{table*}

\clearpage

\begin{table*}[ht!]
\renewcommand{\arraystretch}{1.2}
\setlength{\tabcolsep}{2pt}
    \fontsize{9}{10}\selectfont
    \centering
    \begin{tabular}{p{15cm}}
\toprule
\toprule
{\bf Question:} Which film whose director is younger, My Baby'S Daddy or A Tale Of Winter?
\\\bottomrule
\rowcolor[rgb]{ .900,  .900,  .900}{\textit{"Thought":"}
To determine which film has a director who is younger, we need to identify the directors of both films and compare their ages.}
\\
\rowcolor[rgb]{ .900,  .900,  .900}{First, let's identify the director of "My Baby's Daddy".}
\\
\search{Who directed My Baby's Daddy?}
\\
\hobs \,\, (Title: "My Baby's Daddy") \hlrel{My Baby's Daddy is a 2004 American comedy film, directed by Cheryl Dunye.} 
\\
(Title: "Alec Mapa: Baby Daddy") Alec Mapa: Baby Daddy is a 2013 one- man show by Alec Mapa $\dots$
\\
(Title: "Bachelor Daddy") Bachelor Daddy is a 1941 American comedy film directed by Harold Young and written by Robert Lees and Frederic I. Rinaldo $\dots$
\\
(Title: "Baby and I") Baby and I is a 2008 South Korean film about a rebellious 18- year- old high school senior who ends up caring for a baby with a note $\dots$
\\
(Title: "Who's Your Daddy? (film)") Who's Your Daddy? is a 2002 American comedy film directed( and co-scripted) by Andy Fickman.
\hobs
\\
\rowcolor[rgb]{ .900,  .900,  .900}{The director of "My Baby's Daddy" is Cheryl Dunye.}
\\
\rowcolor[rgb]{ .900,  .900,  .900}{Next, let's identify the director of "A Tale Of Winter".}
\\
\search{Who directed A Tale Of Winter?}
\\
\hobs \,\,
(Title: "A Tale of Winter") \hlrel{A Tale of Winter is a 1992 French drama film directed by Eric Rohmer,} and starring Charlotte Very, Frederic van den Driessche and Michael Voletti.  It is the second of Rohmer's "Tales of the Four SeasonsContes des quatre saisons"), which also include "A Tale of Springtime" (1990), "A Summer's Tale" (1996) and "Autumn Tale" (1998) $\dots$
\\
(Title: "A Tale of Winter") \hlrel{A Tale of Winter is a 1992 French drama film directed by Eric Rohmer,} and starring Charlotte Very, Frederic van den Driessche and Michael Voletti.  It is the second of Rohmer's" Tales of the Four SeasonsContes des quatre saisons") $\dots$
\\
$\dots$
\hobs
\\
\rowcolor[rgb]{ .900,  .900,  .900}{The director of "A Tale of Winter" is Eric Rohmer.}
\\
\rowcolor[rgb]{ .900,  .900,  .900}{
Now, we need to determine the ages of Cheryl Dunye and Eric Rohmer to compare.}
\\
\search{When was Cheryl Dunye born?}
\\
\hobs \,\,
(Title: "Cheryl Dunye") \hlrel{Cheryl Dunye( born May 13, 1966) is a Liberian- American film director, producer, screenwriter, editor and actress.} Dunye's work often concerns themes of race, sexuality, and gender, particularly issues relating to black lesbians. $\dots$
\\
$\dots$
\\
\rowcolor[rgb]{ .900,  .900,  .900}{Cheryl Dunye was born on May 13, 1966.}
\\
\search{When was Eric Rohmer born?}
\\
\hobs \,\,
(Title: "Eric Rohmer") Jean Marie Maurice Scherer or Maurice Henri Joseph Scherer, known as \hlrel{Eric Rohmer (21 March 192011 January 2010),} was a French film director, film critic, journalist, novelist, screenwriter, and teacher. Rohmer was the last of the post-World War II French New Wave directors to become established. He edited the influential film journal, "Cahiers du cinema", from 1957 to 1963, while most of his  $\dots$
\\
(Title: "Eric Rohmer filmography") This is a list of films by the French director Eric Rohmer.
\\
$\dots$
\hobs
\\

\rowcolor[rgb]{ .900,  .900,  .900}{Eric Rohmer was born on March 21, 1920.}
\\
\rowcolor[rgb]{ .900,  .900,  .900}{Now we compare the birth years:}
\\
\rowcolor[rgb]{ .900,  .900,  .900}{- Cheryl Dunye: 1966}
\\
\rowcolor[rgb]{ .900,  .900,  .900}{- Eric Rohmer: 1920}
\\
\rowcolor[rgb]{ .900,  .900,  .900}{Cheryl Dunye is younger than Eric Rohmer.}
\\
\rowcolor[rgb]{ .900,  .900,  .900}{Therefore, the film "My Baby's Daddy" has a younger director.}
\\
\hevi- "My Baby's Daddy" was directed by Cheryl Dunye, born in 1966.
\\
- "A Tale of Winter" was directed by Eric Rohmer, born in 1920.
\\
- Cheryl Dunye is younger than Eric Rohmer.
\hevi
\\
\answer{My Baby'S Daddy} \,\, "\textcolor{green}{\ding{51}}"
\\\bottomrule
\bottomrule
 \end{tabular}
    \caption{{\bf An 4-hop example of R-Search on 2WikiMQA dataset using GRPO.
}
\hlrel{Orange segments} denote those used to support correct answers.
% \hlred{Red segments} represent key incorrect information and erroneous responses collected.
The "\textcolor{green}{\ding{51}}" indicates a correct answer.
}
\label{tab-app: case grpo-2wiki-4hop}
\end{table*}

\end{document}